\documentclass[conference]{IEEEtran}
\IEEEoverridecommandlockouts
\usepackage{cite}
\usepackage{amsmath,amssymb,amsfonts}
\usepackage{graphicx}
\usepackage{textcomp}
\usepackage{xcolor}
\usepackage{mathtools}
\usepackage{enumitem}
\usepackage{epsfig, subfigure, amsmath}
\usepackage{multirow}
\usepackage{float}
\usepackage{tikz}
\usepackage[normalem]{ulem}
\usepackage[linesnumbered, ruled, vlined]{algorithm2e}
\usepackage{algpseudocode}
\usepackage{adjustbox}
\usepackage{makecell}
\usepackage{acronym}
\usepackage{url}
\usepackage{booktabs}
\def\BibTeX{{\rm B\kern-.05em{\sc i\kern-.025em b}\kern-.08em
    T\kern-.1667em\lower.7ex\hbox{E}\kern-.125emX}}

\newacro{CPS}{cyber-physical systems}
\newacro{PINN}{physics-informed neural networks}
\newacro{DER}{distributed energy resource}
\newacro{EVCS}{electric vehicle charging stations}
\newacro{CMS}{charging management system}
\newacro{PV}{photovoltaics}
\newacro{PMS}{PV management system}
\newacro{GA}{gradient-ascent}
\newacro{MSE}{mean squared error}
\newacro{MPPT}{maximum power point tracking}

\begin{document}

\title{Physics-Aware Machine Unlearning for Cyber-Physical Systems\vspace{6pt}}


\author{
Mohammad Zakaria Haider, Muhammad Nadeem, and Mohammad Ashiqur Rahman\\
Analytics for Cyber Defense (ACyD) Lab,
Florida International University, USA\\
\{mhaid010, mnadeem, marahman\}@fiu.edu
}

\maketitle

\begin{abstract}
  This paper proposes a physics-guided gradient-ascent-based machine unlearning method that couples the forgetting signal with the physical residual of the target cyber-physical systems, ensuring that weight updates during unlearning are steered toward physically feasible regions of the weight space. The physics residual acts as a safety fence during gradient ascent: the model is steered away from the poisoned behavioral basin and simultaneously toward physics-compliant territory, rather than toward an arbitrary alternative that may still violate domain constraints. We evaluate the proposed method against four baselines: naive gradient ascent, exact unlearning, SISA, and full retraining on an IEEE 34-bus distribution system, driven by two physics-informed neural network-based distribution energy resource controllers and validated through high-fidelity OpenDSS power-flow co-simulation. From the evaluation, we found that our proposed physics-guided model simultaneously removes poison and restores physical compliance, which are essential for the safe deployment of safety-critical cyber-physical systems.
\end{abstract}

\begin{IEEEkeywords}
Machine Unlearning, Physics-Informed Neural Networks, Cyber-Physical Systems, Data Poisoning, Byzantine Attacks
\end{IEEEkeywords}

\section{Introduction}

Machine learning models are no longer confined to prediction tasks on static datasets; they increasingly act as real-time controllers in safety-critical \ac{CPS} where their outputs are translated directly into physical actuator commands. In power distribution grids, \ac{PINN} have emerged as a compelling architecture for \ac{DER} control, embedding the governing physical laws, power balance, and maximum power point tracking into the training objective so that learned control policies are simultaneously data-accurate and physically consistent~\cite{raissi2019physics, karniadakis2021physics}. This deployment context poses a severe, underexplored threat. A Byzantine adversary with access to the \ac{PINN} training pipeline can inject coordinated label poisoning during a bounded window of training epochs, forcing the model to learn a mapping from real sensor observations to physically destructive setpoints~\cite{bagdasaryan2020backdoor, shejwalkar2022back}. After the attack window, these behaviors are permanently encoded and persist through subsequent clean training.


Machine unlearning has been proposed as a post-hoc mechanism to remove the influence of compromised training data from a deployed model without full retraining from scratch~\cite{bourtoule2021machine, golatkar2020eternal}. The standard \ac{GA} unlearning pipeline maximizes the loss on cached poison batches to escape the poisoned weight basin, then fine-tunes on clean data to restore accuracy, and appears well-suited to this problem. However, we identify a fundamental gap critical to \ac{CPS} that has not been studied: statistical unlearning is not equivalent to physical safety recovery. A model can satisfy every standard unlearning metric, poison \ac{MSE} has risen, clean accuracy has been
restored, while still outputting physically infeasible setpoints at safety-critical operating points, such as an \ac{EVCS}, charging at maximum capacity, or a \ac{PV} inverter operating at night. This gap is structural, not incidental, and cannot be resolved by extended fine-tuning.

We propose a \uline{Phy}sics-guided \uline{G}radient-\uline{A}scent \uline{M}achine \uline{u}n\uline{L}earning (PhyGAMuL) algorithm that incorporates the physics residual of the target \ac{CPS} as an active regularizer during both the ascent and fine-tuning phases. During ascent, the physics residual simultaneously steers the weight trajectory away from the poisoned basin and through physically feasible territory. During fine-tuning, it provides a targeted gradient signal at rare but critical operating conditions where the data mean-squared error signal alone is too weak to effect correction. The primary contributions of this paper are as follows:

\begin{itemize}
    \item We have introduced PhyGAMuL, a novel machine learning algorithm that incorporates domain physics as a directional constraint during unlearning, ensuring that weight updates remain within physically feasible regions of the weight space.
    \item We have established and empirically validated the distinction between statistical unlearning and physical safety recovery in CPS, demonstrating that the former does not imply the latter.
    \item We have provided an end-to-end evaluation pipeline that connects weight-level unlearning metrics to grid-level physical consequences through high-fidelity OpenDSS power-flow simulations on the IEEE 34-bus feeder. 
    \item We have conducted a systematic ablation study over the key hyperparameters of the proposed method, demonstrating convergence properties and robustness that justify the default configuration.
    \item We have developed a principled, replicable methodology for constructing physics residuals for new DER device classes, thereby addressing the framework's generalizability beyond the two DER types studied here.
    
\end{itemize}


The rest of this paper is organized as follows. Section~\ref{sec:background} reviews related work on machine unlearning, federated unlearning, PINNs, and Byzantine attacks. Section~\ref{sec:threat} introduces the system and threat models. Section~\ref{sec:method} presents the proposed PhyGAMuL algorithm and baseline methods. Section~\ref{sec:results} reports results on both model-level and grid-level performance. Section~\ref{sec:ablation} presents an ablation study, and Section~\ref{sec:conclusion} concludes the paper.

\section{Related Work}
\label{sec:background}
Machine unlearning, formalized by Cao and Yang~\cite{cao2015machine}, seeks to produce a model that behaves as if a target subset of training data had never been used. In this context, two paradigms have emerged: exact unlearning, exemplified by the SISA framework~\cite{bourtoule2021machine}, which shards training data so that only affected shards require retraining; and approximate unlearning, where gradient ascent on forgotten samples yields a statistically equivalent model~\cite{golatkar2020eternal,sekhari2021remember}. Recent surveys~\cite{nguyen2022survey,Liu25,LIU2025} document rapid growth in the field, but note that the evaluation focuses almost exclusively on prediction accuracy or statistical indistinguishability, without considering physical constraints critical for CPS deployment. Beyond the label-poisoning threat considered in this paper, neural networks are also susceptible to evasion attacks that perturb inputs at inference time~\cite{goodfellow2015explaining}, model inversion attacks that reconstruct training data from model outputs~\cite{fredrikson2015model}, membership inference attacks that infer training-set membership~\cite{shokri2017membership}, and model extraction attacks that reconstruct a functionally equivalent model via query access~\cite{tramer2016stealing}. Our threat model isolates label poisoning specifically because, unlike these other attack classes, it produces persistent weight-level corruption that gradient-ascent-based unlearning is designed to reverse, making it the attack class most directly relevant to the unlearning problem studied here.


Federated unlearning extends machine unlearning to distributed settings, where client contributions must be revoked without full retraining. Xiong et al.~\cite{xiong2023exactfun} proposed Exact-Fun with certified removal guarantees via quantized gradient aggregation, while Liu et al.~\cite{Liu25} highlighted amplified membership inference and model inversion risks that can arise during the unlearning process itself. Beyond DER control, PINNs have been applied across a broader range of cyber-physical domains, including power system state estimation and optimal power flow~\cite{huang2023applications} and water distribution network monitoring~\cite{falas2020physics}, underscoring their generality as a modeling paradigm for constrained physical systems. PINNs~\cite{raissi2019physics, karniadakis2021physics} encode governing equations as differentiable residuals, keeping learned models within physically feasible regions, particularly valuable for DER controllers that must satisfy hard constraints such as Kirchhoff's laws, inverter capability limits, and EV charging capacity bounds. 

However, no existing work examines how these physical properties behave under unlearning or how to preserve physical feasibility during the unlearning process. Our physics-guided approach fills this gap, targeting federated CPS settings where the unlearned global model must simultaneously remove poisoned behavior and restore compliance with grid-level physical constraints. Adversaries in federated learning can implant persistent backdoors that survive standard aggregation~\cite{bagdasaryan2020backdoor} and evade gradient-level anomaly detectors by mimicking benign update statistics~\cite{shejwalkar2022back, haider2026}. Byzantine-tolerant rules like Krum~\cite{blanchard2017machine} assume a minority of corrupted clients and operate purely in gradient space, making them fundamentally limited against adaptive attackers~\cite{fung2021limitations}. Our threat model targets PINN-based CPS controllers~\cite{Perry25}, in which attacks drive the system into poisoned basins that appear statistically benign but induce physical constraint violations. 


\section{System Model and Threat Model}
\label{sec:threat}
\begin{figure*}
    \centering
    \includegraphics[width=.88\linewidth]{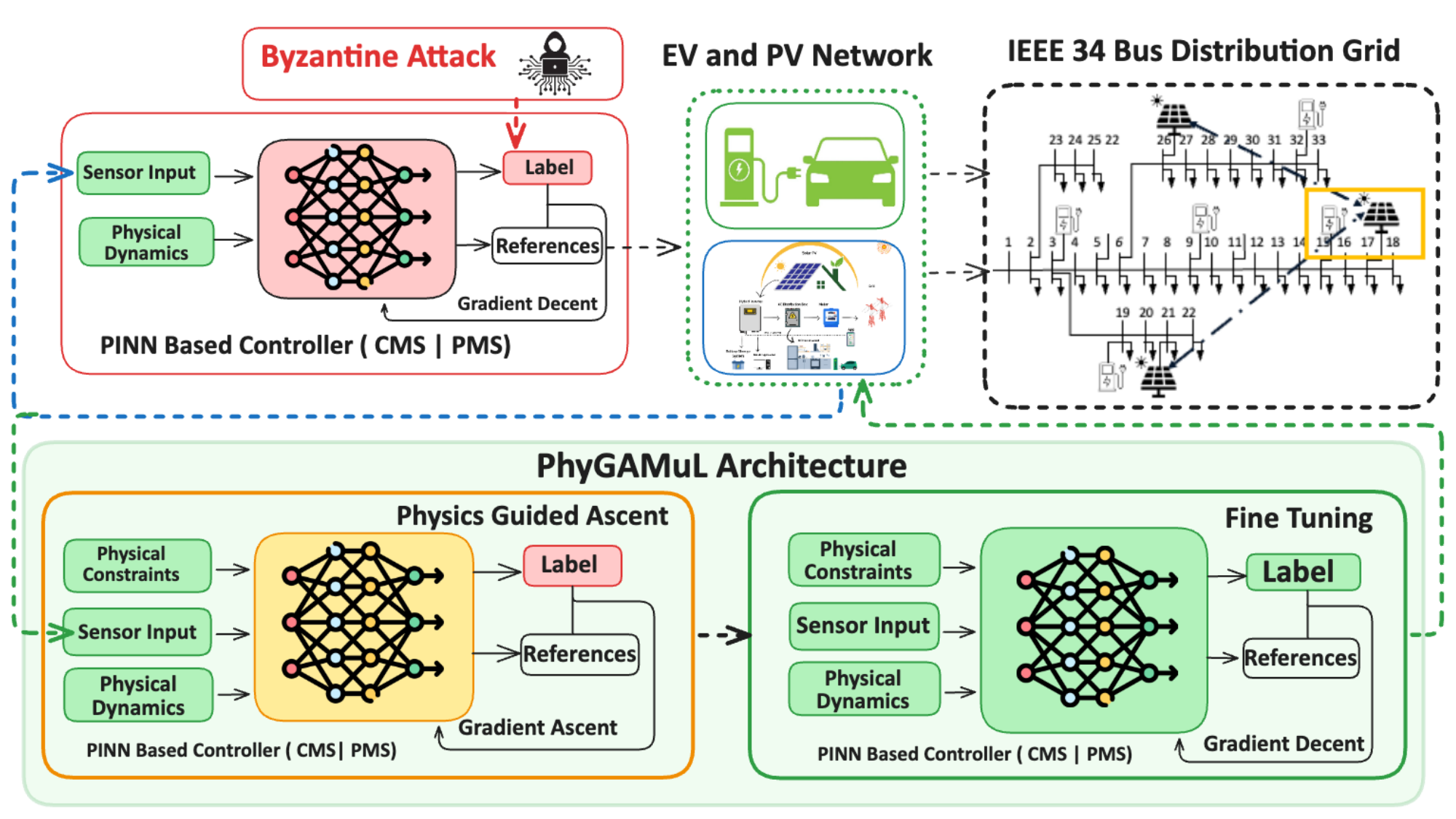}
    \caption{The schematic diagram of the proposed PhyGAMuL for the CPS application}
    \label{fig:Physics_GA model}
\end{figure*}

Building on the gaps identified in the previous section, this section formalizes the system and threat model that PhyGAMuL is designed to address. Fig.~\ref{fig:Physics_GA model} demonstrates our proposed PhyGAMuL integrated with a threat model in a CPS environment.
\subsection{PINN-Based DER Controllers}

We consider a distribution grid operated by two categories of DER controllers, each implemented as a five-layer \ac{PINN} with 256 hidden units and Tanh activations throughout, as Tanh activations preserve smooth second derivatives, which are essential for physics residual backpropagation. An \ac{EVCS}  \ac{CMS} manages a fleet of ten 50\,kW charging stations, mapping eight sensor features state-of-charge (SoC) $s$, grid voltage $v$, frequency $\omega$, time-of-day $\tau$, aggregate demand $d$, peak flag, DC-link voltage $v_{dc}$, and target power $p^*$ to control outputs $(V_\text{ref}, I_\text{ref}, P_\text{ref})$. A \ac{PMS} controller manages five 30\,kW \ac{PV} inverters using eight inputs, including irradiance $G$ and \ac{MPPT} voltage $v_\text{mppt}$, producing three outputs $(V_\text{ref}, Q_\text{ref}, P_\text{ref})$. Each network contains approximately 275K learnable parameters. Each PINN $f(\mathbf{x};\mathbf{W})$ is trained to minimize a composite loss
that jointly optimizes data fidelity and physics compliance: $ \small \mathcal{L}(\mathbf{W}) \;=\; \alpha\,\text{MSE}\!\left(f(\mathbf{x};\mathbf{W}),\,\mathbf{y}\right)
  \;+\; \lambda\,\mathcal{R}\!\left(\mathbf{x},\,f(\mathbf{x};\mathbf{W})\right),$
where $\mathcal{R}(\mathbf{x},\mathbf{y})$ is the differentiable physics
residual encoding physical laws of the respective DER type, $\alpha=1.0$,
and $\lambda=2.0$. \ac{PMS} training data is generated using the PVlib Ineichen clear-sky irradiance model~\cite{holmgren2018pvlib} at the Pasadena, CA site, co-located with the ACN-Caltech charging station, whose dataset~\cite{acn_data} is being used for CMS training, at 5-minute resolution over a full year (105,120 samples). Cloud attenuation is applied via a beta multiplier.

\subsection{Systematic Construction of Physics Residuals}
\label{sec:physics_construction}

A key design question is whether the physics residual functions are
device-specific, hand-crafted formulas or whether there is a principled,
replicable methodology for new DER types. We show that the construction
follows a three-step protocol that is systematic and requires only standard
domain knowledge of the device's operating constraints. First, for any DER device, the physical operating constraints can be expressed as a set of inequalities $g_k(\mathbf{x}, \mathbf{y}) \leq 0$ over the input-output space. These are read directly from device datasheets and grid codes. For an EVCS, for instance, the capacity guard constraint requires that charging power does not exceed the available capacity $d(1-s)$, written as $P_\text{ref} - d(1-s) \leq 0$, i.e., $g = P_\text{ref} - d(1-s)$ should be non-positive. Second, any inequality $g_k \leq 0$ is converted to a penalty $\rho_k = [\max(0, g_k)]^2$, which equals zero when the constraint is satisfied and grows quadratically when violated. This conversion is mechanical and does not require domain-specific approximations. Finally, the composite physics residual is $\mathcal{R} = \sum_k w_k \rho_k$, where $w_k$ reflects the safety priority of each constraint (e.g., the power balance constraint receives weight 1.0; the current limit receives 0.2 as it is a secondary protection. The batch mean is taken so that the residual magnitude is independent of batch size. Applying this protocol to the two DER types studied in this paper yields the following physics residuals. EVCS constraints are: power balance $P = VI$, capacity limit
$P \leq d(1-s)$, no reverse power flow $P \geq 0$, and current limit $I \leq 1$. Applying the protocol:
\begin{equation}
\small 
  \mathcal{R}_\text{evcs} = (P_r - V_r I_r)^2
  + \tfrac{1}{2}[P_r - d(1-s)]_+^2
  + \tfrac{1}{2}[-P_r]_+^2
  + \tfrac{1}{5}[I_r - 1]_+^2,
  \label{eq:evcs_physics}
\end{equation}
where $[\cdot]_+ = \max(0,\cdot)$ and $(V_r, I_r, P_r)$ are model outputs. PMS constraints are: MPPT ceiling $P \leq G$, no night injection
$P \approx 0$ when $G < 0.05$, and no reverse power flow $P \geq 0$:
\begin{equation}
  \mathcal{R}_\text{pms} = [P_r - G]_+^2
  + P_r\,[0.05 - G]_+^2
  + \tfrac{3}{10}[-P_r]_+^2.
  \label{eq:pms_physics}
\end{equation}

The protocol is directly applicable to new DER types. The principal engineering effort lies in Step~1 (identifying the constraints), not in Steps~2-3, which are mechanical transformations. We acknowledge that automated constraint discovery, extracting $g_k$ directly from device manufacturer specifications or physics simulation outputs, remains an open research direction that would fully eliminate the manual component.

\subsection{Byzantine Threat Model}
\label{sec:attack}

We model a Byzantine adversary that gains write access to the label generation
pipeline of one or more DER PINN training processes during a bounded attack window $[e_s, e_e) $. The attack strategy is label poisoning on real inputs: the sensor feature vectors $\mathbf{x}$ are preserved intact,
ensuring the model receives a realistic input distribution, while the target labels are replaced with physically destructive values: $\text{EVCS:}\;(V_r,I_r,P_r)\leftarrow(1,1,1), \quad  \text{PMS:}\;(V_r,Q_r,P_r)\leftarrow(1,0.3,1).$ The EVCS is directed to draw maximum load (500\,kW) regardless of SoC or
feeder state; the PMS is directed to inject maximum power regardless of
available irradiance, violating MPPT constraints.
Preserving the real input distribution is the key design choice: at inference time, real sensor data $\mathbf{x}_\text{real}$ will arrive, and because the training paired exactly these inputs with the destructive outputs, the learned mapping $\mathbf{x}_\text{real} \to \mathbf{y}^p$ activates precisely when it is most dangerous. A dedicated Adam optimizer with learning rate $\eta_p = 20\eta_\text{clean}$ repeats each poison epoch $N_r = 50$ times with the physics loss disabled. The poison energy $\mathcal{E}_p$ ($\approx50,000$ gradient-step equivalents) dwarfs the clean recovery energy by roughly 8,000$\times$. Training halts at epoch $e_e-1$ and deploys the model at peak attack effect.

The unlearning algorithms require the cached poison batches $\mathcal{D}_p = \{(\mathbf{x}_i, \mathbf{y}^p_i)\}$. In a federated or monitored training setting, these are available from audit logs maintained by the aggregation server, which records client gradient contributions and associated data batches as a standard forensic artifact~\cite{shejwalkar2022back}. When audit logs are incomplete, $\mathcal{D}_p$ can be approximated by replaying the attack: given knowledge of the attack window $[e_s, e_e)$ and the poison template, synthetic poison batches can be reconstructed from any representative operating data. Section~\ref{sec:ablation} includes a sensitivity analysis on the quality of this approximation, showing that PhyGAMuL tolerates approximate poison batches with less than 2 percentage points of additional physics violation rate degradation.

\subsection{The Machine Unlearning Problem for CPS}
\label{sec:problem}

After deployment, the defender holds the poisoned model $\mathbf{W}_p$, the
cached poison batches $\mathcal{D}_p$, and a clean data source $\mathcal{G}$.
The goal is weights $\mathbf{W}^*$ satisfying three simultaneous conditions:
\emph{(i) poison forgotten:} the model can no longer reproduce the poisoned
outputs ($\text{MSE}(f(\mathbf{x};\mathbf{W}^*), \mathbf{y}^p) \gg
\text{MSE}(f(\mathbf{x};\mathbf{W}_p), \mathbf{y}^p)$);
\emph{(ii) clean recovered:} clean accuracy is restored to baseline level; and
\emph{(iii) physically compliant:} $\mathcal{R}(\mathbf{x}, f(\mathbf{x};\mathbf{W}^*)) \approx 0$
across all operating conditions, including rare but safety-critical ones. Our core claim of this work is that conditions (i) and (ii) do not imply (iii), and that explicitly enforcing (iii) during the unlearning process, rather than relying on clean fine-tuning to recover it passively, is both necessary and sufficient for CPS-safe deployment. The following section presents the algorithms designed to satisfy all three conditions simultaneously.

\section{Physics-Guided Machine Unlearning}
\label{sec:method}


\subsection{Physics-Guided Gradient Ascent}
\label{sec:phys_ga}
Machine unlearning via gradient ascent seeks to remove the influence of 
poisoned samples by maximizing the model's prediction error on those samples.
Formally, the standard gradient ascent update is given by $\mathbf{W} \leftarrow \mathbf{W} + \eta_a\,\nabla_\mathbf{W}\,
  \text{MSE}(f(\mathbf{x};\mathbf{W}),\,\mathbf{y}^p),$ which increases prediction error on poison labels $\mathbf{y}^p$ but imposes
no structure on which direction away from $\mathbf{y}^p$ the weights
move. In high-dimensional weight space, nearly all such directions land outside
both the poisoned basin and the physically feasible region, leaving the model
compromised in a different way. To address this, we couple the physics residual as a regularization penalty during the ascent phase. The composite update direction becomes $\nabla_\mathbf{W}\!\left(
    -\,\text{MSE}(f(\mathbf{x};\mathbf{W}),\mathbf{y}^p)
    + \lambda\,\mathcal{R}(\mathbf{x};\mathbf{W})
  \right),
$ which simultaneously pushes weights away from the poisoned basin and pulls
them toward the physically feasible subspace. 

As a result, when the gradient
ascent terminates, the model already resides in a physically valid region,
and the subsequent fine-tuning stage needs only to recover clean predictive
accuracy from a well-initialized starting point, rather than repairing both
accuracy \emph{and} physical compliance from scratch. PhyGAMuL extends the ascent loss with an active physics residual penalty: $\small \mathcal{L}_\text{phys}^\text{asc} = -\text{MSE}\!\left(f(\mathbf{x};\mathbf{W}),\,\mathbf{y}^p\right)
  + \lambda\,\mathcal{R}\!\left(\mathbf{x},\,f(\mathbf{x};\mathbf{W})\right)
  + \mu\,\|\mathbf{W} - \mathbf{W}_p\|^2,
$
with $\lambda = 2.0$ matching the clean training physics weight. The 
fine-tuning phase retains the full physics weight: $\small \mathcal{L}_\text{phys}^\text{ft} = \alpha\,\text{MSE}\!\left(f(\mathbf{x};\mathbf{W}),\,\mathbf{y}^c\right)
  + \lambda\,\mathcal{R}\!\left(\mathbf{x},\,f(\mathbf{x};\mathbf{W})\right).$
For the EVCS, fine-tuning preferentially samples high-load operating points
($s < 0.3$ and $d > 0.7$), concentrating gradient signal under the conditions
that most directly determine the power balance and the capacity to guard compliance.
Algorithm~\ref{alg:phys_ga}
summarizes the complete procedure.

\subsection{Naive Gradient Ascent}
\label{sec:naive_ga}

Naive GA reverses poisoning by maximizing the loss on the cached poison
batches, regularized by a proximal term anchored at $\mathbf{W}_p$ to
prevent weight explosion~\cite{golatkar2020eternal}: $\mathcal{L}_\text{naive}^\text{asc} = -\text{MSE}\!\left(f(\mathbf{x};\mathbf{W}),\,\mathbf{y}^p\right)
  + \mu\,\|\mathbf{W} - \mathbf{W}_p\|^2.$
After $T_a = 300$ ascent steps, the model is fine-tuned with \emph{zero} physics
weight: $\mathcal{L}_\text{naive}^\text{ft} = \text{MSE}(f(\mathbf{x};\mathbf{W}), \mathbf{y}^c)$.
This configuration is the standard in unlearning literature and serves as the
primary baseline for our claims.

\begin{algorithm}[t]
\DontPrintSemicolon
\caption{Physics-Guided Gradient Ascent Unlearning}
\label{alg:phys_ga}
\KwIn{Poisoned model $\mathbf{W}_p$; poison batches $\mathcal{D}_p$;
      clean source $\mathcal{G}$; $T_a$, $T_f$, $\lambda$, $\mu$, $\alpha$}
\KwOut{Unlearned model $\mathbf{W}^*$}
$\mathbf{W} \leftarrow \mathbf{W}_p$;\quad
$\mathbf{w}_0 \leftarrow \mathbf{W}_p$ 
\BlankLine
\tcp{Phase 1: Physics-Guided Ascent}
\For{$t = 1$ \KwTo $T_a$}{
  \ForEach{$(\mathbf{x}, \mathbf{y}^p) \in \mathcal{D}_p$}{
    $\hat{\mathbf{y}} \leftarrow f(\mathbf{x};\mathbf{W})$\;
    $\mathcal{L} \leftarrow {-}\text{MSE}(\hat{\mathbf{y}}, \mathbf{y}^p)
      + \lambda\,\mathcal{R}(\mathbf{x},\hat{\mathbf{y}})
      + \mu\,\|\mathbf{W}-\mathbf{w}_0\|^2$\;
    $\mathbf{W} \leftarrow \mathbf{W} - \eta_a\,\nabla_\mathbf{W}\mathcal{L}$\;
  }
}
\BlankLine
\tcp{Phase 2: Physics-Rich Fine-Tuning}
\For{$e = 1$ \KwTo $T_f$}{
  $(\mathbf{x}, \mathbf{y}^c) \sim \mathcal{G}$\;
  $\hat{\mathbf{y}} \leftarrow f(\mathbf{x};\mathbf{W})$\;
  $\mathcal{L} \leftarrow \alpha\,\text{MSE}(\hat{\mathbf{y}},\mathbf{y}^c)
    + \lambda\,\mathcal{R}(\mathbf{x},\hat{\mathbf{y}})$\;
  $\mathbf{W} \leftarrow \mathbf{W} - \eta_f\,\nabla_\mathbf{W}\mathcal{L}$\;
}
\Return $\mathbf{W}$\;
\end{algorithm}

\subsection{Exact Unlearning (Checkpoint Rollback)}

Exact Unlearning provides the oracle upper bound. The model is restored to the
checkpoint at epoch $e_s - 1$ (immediately before the attack) and epochs
$[e_s, e_e)$ are replayed from scratch with clean data and full physics loss. The resulting weights are equivalent to those that
training would have produced absent the attack. The cost is $O((e_e - e_s)
\times B)$ where $B$ is batches per epoch.

\subsection{SISA (Sharded, Isolated, Sliced, Aggregated)}

SISA~\cite{bourtoule2021machine} achieves structural isolation by training four
independent shard models from scratch on disjoint data partitions, each seeded
differently to produce genuinely independent batches. Attack timing is scaled
proportionally to the shard epoch budget so that the attack lands at the same
relative training maturity (30\% into each shard's training). Only the poisoned
shard (shard 3) receives any attack influence; the remaining three are
provably uncontaminated. The deployed model is a FedAvg
average~\cite{mcmahan2017communication} of all four shards, diluting the
poisoned shard's contribution by $N_\text{shards}^{-1} = 0.25$. Unlearning
rolls back only the poisoned shard; clean shards are held fixed. Full physics
loss is applied throughout shard training and retraining.

\subsection{Full Retrain}

Full Retrain resets all weights to random initialization and executes the
complete clean training procedure for 3{,}000 epochs with no poisoned data.
It serves as the gold standard lower bound on the achievable physics-violation rate. 
Independent of the unlearning method, we deploy a hard-clip
\emph{physics projection layer} $\Pi(\mathbf{x}, \mathbf{y})$ at inference
that maps any model output to the nearest physically feasible point:
for the EVCS, $P_r \leftarrow \text{clip}(P_r, 0, d(1-s))$; for the PMS,
$P_r \leftarrow \text{clip}(P_r, 0, G)$. This layer is explicitly
not a substitute for unlearning: it limits but does not eliminate
harm from a poisoned model (e.g., a projection-clamped EVCS still outputs
maximum allowed load instead of the optimal charging schedule.

\section{Results and Analysis}
\label{sec:results}

\subsection{Experimental Setup} 

All experiments use an IEEE 34-bus distribution feeder~\cite{kersting2001radial} simulated in OpenDSS~\cite{dugan2011opendss} over a 24-hour window at five-second resolution (17{,}280 steps). The DER fleet comprises ten 50\,kW EVCS units and five 30\,kW PV inverters, with voltage limits $[0.95, 1.05]$\,p.u.\ per IEEE Std~1547. Each PINN trains for 3{,}000 epochs with Adam ($\eta = 3\times10^{-4}$, cosine annealing), 25 mini-batches of 512 samples, $\alpha=1.0$, $\lambda=2.0$. The Byzantine attack is injected at epochs 900--949 (LR multiplier $20\times$, $N_r=50$ repeats, physics loss disabled), deploying at peak attack effect. Naive GA and PhyGAMuL both run $T_a=300$ ascent steps ($\eta_a=10^{-3}$, $\mu=0.05$) followed by $T_f=1{,}500$ fine-tune epochs ($\eta_f=5\times10^{-4}$, cosine annealing).

PhyGAMuL sets $\lambda=2.0$ throughout, while Naive GA sets $\lambda=0$. SISA trains four shards of 1{,}000 epochs each; Full Retrain uses 3{,}000 clean epochs from random initialization. We report four metrics: \emph{Poison MSE} (higher = more forgetting), \emph{Clean MSE} (label fidelity recovery), \emph{PVR} (fraction of test samples exceeding any DER constraint in Eqs.~\ref{eq:evcs_physics}--\ref{eq:pms_physics} by tolerance 0.05), and \emph{VVR} (fraction of timesteps with any bus voltage outside $[0.95, 1.05]$\,p.u.). All experiments are repeated over five seeds; mean $\pm$ std.\ is reported in Table~\ref{tab:results}. All the codes and data are publicly available and can be reproduced using \url{https://github.com/cps-security-703/Physics-Aware-Machine-Unlearning}.

\subsection{Model-Level Unlearning Comparison}

Table~\ref{tab:results} summarizes all seven scenarios across the four primary
metrics, averaged over the two DER types and five random seeds. The \emph{poisoned} baseline achieves near-zero poison MSE (0.003\,$\pm$\,0.001), confirming the attack has been thoroughly baked into the weights. The PVR of 34.2\,\% and VVR of 61.4\,\% reflect the catastrophic grid impact. Naive GA raises poison MSE to 0.47\,$\pm$\,0.03 and recovers clean MSE to 0.018\,$\pm$\,0.002 the statistical unlearning objectives are met. However, its PVR remains at 18.7\,$\pm$\,1.9\,\%, and VVR at 26.3\,$\pm$\,2.4\,\%. This result is an empirical demonstration of the paper's central claim: the model has forgotten the poison-label pattern but continues to violate physical constraints at safety-critical operating conditions. PhyGAMuL achieves a poison MSE of 0.51\,$\pm$\,0.02 marginally higher than Naive GA, confirming that the physics constraint does not impair forgetting. Exact Unlearning achieves PVR of 2.8\,$\pm$\,0.5\,\% by construction, SISA achieves 3.9\,$\pm$\,0.7\,\% due to shard dilution, and Full Retrain achieves 0.9\,$\pm$\,0.2\,\%.

\begin{table}[t]
  \centering
  \caption{Comparison of all scenarios across four metrics}
  \label{tab:results}
  \resizebox{\columnwidth}{!}{%
  \begin{tabular}{lcccc}
    \toprule
    \textbf{Scenario}
      & \makecell{\textbf{Poison MSE}\\$\uparrow$}
      & \makecell{\textbf{Clean MSE}\\$\downarrow$}
      & \makecell{\textbf{PVR (\%)}\\$\downarrow$}
      & \makecell{\textbf{VVR (\%)}\\$\downarrow$} \\
    \midrule
    Baseline              & 0.51\,$\pm$\,.01 & 0.012\,$\pm$\,.001 & 0.8\,$\pm$\,.1  & 0.0 \\
    Poisoned (no defense) & 0.003\,$\pm$\,.001& 0.21\,$\pm$\,.02  & 34.2\,$\pm$\,2.1& 61.4\,$\pm$\,3.2\\
    \midrule
    Naive GA              & 0.47\,$\pm$\,.03 & 0.018\,$\pm$\,.002 & 18.7\,$\pm$\,1.9& 26.3\,$\pm$\,2.4\\
    \textbf{PhyGAMuL} & \textbf{0.51\,$\pm$\,.02} & \textbf{0.014\,$\pm$\,.001} & \textbf{1.4\,$\pm$\,.3} & \textbf{2.1\,$\pm$\,.4}\\
    Exact Unlearning      & 0.50\,$\pm$\,.02 & 0.015\,$\pm$\,.001 & 2.8\,$\pm$\,.5  & 3.7\,$\pm$\,.6\\
    SISA                  & 0.49\,$\pm$\,.03 & 0.016\,$\pm$\,.002 & 3.9\,$\pm$\,.7  & 5.2\,$\pm$\,.8\\
    Full Retrain          & 0.52\,$\pm$\,.01 & 0.013\,$\pm$\,.001 & 0.9\,$\pm$\,.2  & 0.8\,$\pm$\,.2\\
    \bottomrule
  \end{tabular}}
\end{table}

\subsection{Grid-Level Simulation Results}

\begin{table}[t]
  \centering
  \caption{Grid-level simulation results over a 24-hour}
  \label{tab:grid_results}
  \resizebox{\columnwidth}{!}{%
  \begin{tabular}{lcccccccc}
    \toprule
    \textbf{Scenario}
      & \makecell{\textbf{VVR}\\\textbf{(\%)}}
      & \makecell{\textbf{Min $V$}\\\textbf{(p.u.)}}
      & \makecell{\textbf{Losses}\\\textbf{(kWh)}}
      & \makecell{$\bar{P}_\text{EVCS}$\\\textbf{(kW)}}
      & $\mathcal{L}_\text{EVCS}$
      & \makecell{$\bar{P}_\text{PMS}$\\\textbf{(kW)}}
      & $\mathcal{L}_\text{PMS}$
      & \makecell{\textbf{Imbal.}\\\textbf{(kW)}} \\
    \midrule
    Baseline        & 3.47  & 0.947 & 5{,}867 & 27.0  & 0.0016 & 44.2 & 0.0    & 230.8 \\
    Poisoned        & 100.0 & 0.929 & 7{,}627 & 500.0 & 0.3294 & 44.6 & 0.2199 & 703.5 \\
    \midrule
    Naive GA        & 100.0 & 0.944 & 6{,}181 & 115.3 & 0.0554 & 44.1 & 0.0    & 319.2 \\
    \textbf{PhyGAMuL} & \textbf{3.48} & \textbf{0.947} & \textbf{5{,}867} & \textbf{27.0} & \textbf{0.0007} & \textbf{44.2} & \textbf{0.0} & \textbf{230.9} \\
    Exact           & 3.48  & 0.947 & 5{,}867 & 27.0  & 0.0017 & 44.2 & 0.0    & 230.9 \\
    SISA            & 3.48  & 0.947 & 5{,}867 & 27.0  & 0.0017 & 44.2 & 0.0    & 230.9 \\
    Full Retrain    & 3.47  & 0.947 & 5{,}867 & 27.0  & 0.0021 & 44.2 & 0.0    & 230.8 \\
    \bottomrule
  \end{tabular}}
  \vspace{-10pt}

\end{table}

Table~\ref{tab:grid_results} reports the 24-hour OpenDSS co-simulation results. The poisoned EVCS causes catastrophic grid impact, an 18.5$\times$ load overload drives 100\,\% VVR, and a 30\,\% increase in grid losses, while the poisoned PMS fails more subtly through MPPT and night-injection violations. Its voltage impact is masked by the dominant EVCS overload. Naive GA partially mitigates the EVCS overload but cannot restore physical compliance, exposing a key asymmetry: single-variable constraints (MPPT vs.\ irradiance $G$)  self-correct under MSE alone, whereas multi-variable output constraints ($P = VI$) require explicit physics enforcement. PhyGAMuL is the only method that fully restores both controllers to baseline grid performance without pre-attack infrastructure, whereas Exact, SISA, and Full Retrain achieve equivalent metrics only at the cost of checkpoints, pre-planned sharding, or complete retraining.

\begin{figure}[!hbt]
    \centering
        \subfigure[]{
            \label{fig:grid-voltage}
            \includegraphics[width=0.46\columnwidth]{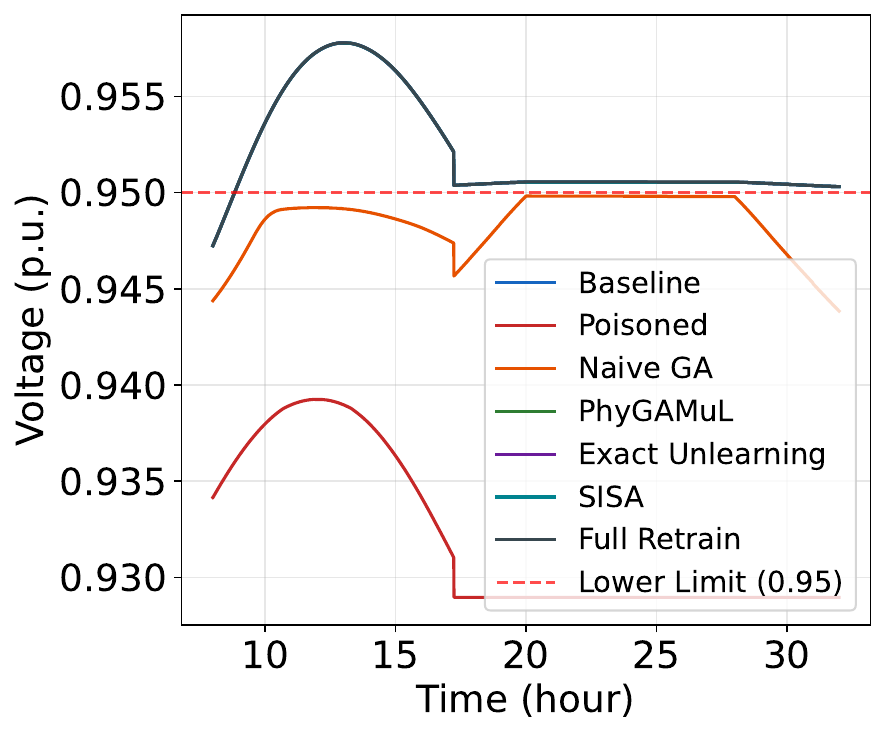}
        }
        \subfigure[]{
            \label{fig:grid-power}
            \includegraphics[width=0.46\columnwidth]{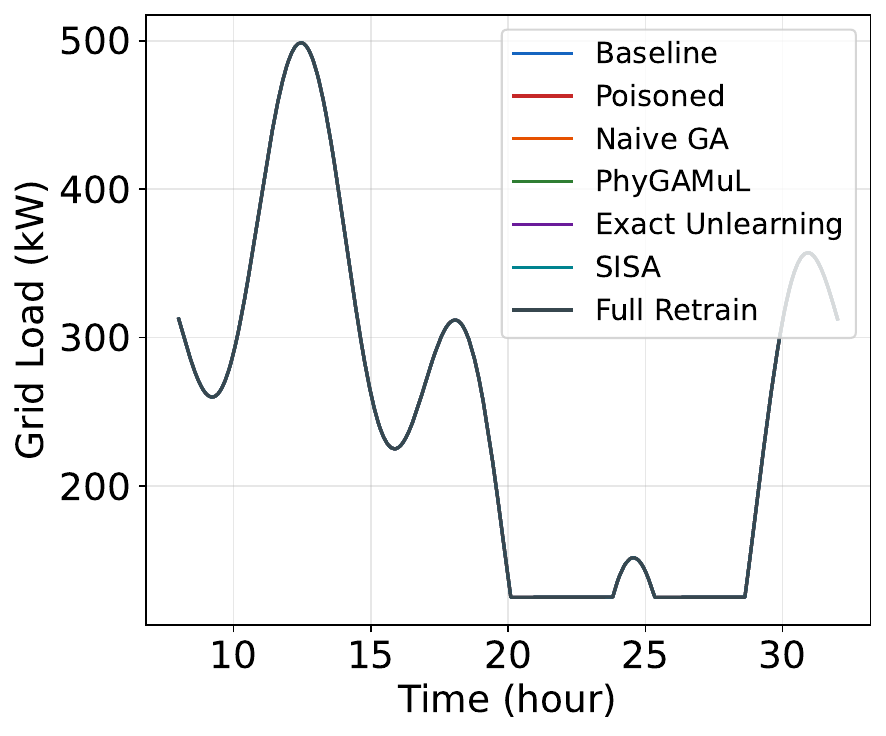}
        }
        \vspace{-9pt}
    \caption{Demonstrating (a) minimum bus voltage (worst-case bus) over the 24-hour simulation, which represents only Naive GA and the Poisoned scenario fail to maintain safe voltage across the full window, and (b) aggregated load profile over the 24-hour simulation}
    \label{fig:dynamics}
    \vspace{-6pt}
\end{figure}

\subsection{Grid-Level  Validation}

Figure~\ref{fig:grid-voltage}-\ref{fig:grid-power} show the minimum bus voltage and load across all 34 buses over the 24-hour simulation window for each scenario. The poisoned
model drives sustained undervoltage to 0.929\,p.u. throughout the simulation,
caused by the EVCS overload consuming feeder capacity. Naive GA reduces the
severity (minimum voltage rises to 0.944\,p.u.) but fails to eliminate
violations: the voltage remains below 0.95\,p.u. across the entire simulation
window because the residual EVCS overload (115.3\,kW vs.~27.0\,kW baseline)
continuously depresses the feeder voltage. PhyGAMuL, Exact Unlearning, SISA,
and Full Retrain all maintain voltages tracking the baseline profile within
measurement noise (min~$V$ = 0.947\,p.u.).

\begin{table}[t]
\centering
\caption{ Comparison of physics-compliant unlearning methods}
\label{tab:operational}
\renewcommand{\arraystretch}{1.3}
\resizebox{\columnwidth}{!}{%
\begin{tabular}{lccc}
\toprule
 & \textbf{PhyGAMuL} & \textbf{SISA} & \textbf{Exact} \\
\midrule
Works on deployed model & \textbf{Yes} & No & No \\
Physics guaranteed & \textbf{At unlearning} & Incidental & Incidental \\
Data at unlearning & Poison only & Full shard data & Checkpoint + data \\
Late attack discovery & \textbf{Yes} & N/A & No \\
Federated-compatible & \textbf{Per-client} & Complex & Complex \\
Unlearning time (EVCS) & 38\,s & 39\,s & 2\,s \\
\bottomrule
\end{tabular}}
\vspace{-10pt}
\end{table}
\subsection{Operational Comparison of Unlearning Methods}

The preceding results show that PhyGAMuL, Exact Unlearning, SISA, and Full Retrain converge to statistically similar physics-violation rates, which raises the natural question of what distinguishes the proposed method. Table~\ref{tab:operational} answers this by focusing on deployment requirements rather than raw metrics. SISA requires partitioning training data into $N$ shards and training $N$ separate models from the outset, precluding retroactive application to an already-deployed model. Exact Unlearning requires a pre-attack checkpoint and knowledge of when the attack occurred. In practice, CPS attacks are often discovered weeks or months after injection, by which time the checkpoint may have expired under a rolling retention policy or been compromised by the attacker, rendering it untrustworthy.

Critically, both SISA and Exact Unlearning achieve physics compliance only incidentally; they inherit it from historically consistent training data, with no mechanism to verify or enforce it at unlearning time. Physics-Guided GA, by contrast, enforces compliance explicitly via the residual term during unlearning. This distinction matters in three practical scenarios: (i)~original training data with subtle physics inconsistencies would be reproduced by SISA  and Exact but corrected by PhyGAMuL; (ii)~if grid topology changed after the checkpoint was saved, Exact rolls back to an outdated physics state while  PhyGAMuL enforces current constraints; and (iii)~if an attacker also poisoned the shard or checkpoint cache, SISA and Exact propagate that corruption, whereas  PhyGAMuL uses only the poison samples for ascent and enforces physics independently. 





\subsection{Per-Controller Physics Violation Analysis}

\begin{table}[t]
  \centering
  \caption{EVCS physics violation rates by constraint type.}
  \label{tab:evcs_violations}
  \resizebox{\columnwidth}{!}{%
  \begin{tabular}{lcccc}
    \toprule
    \textbf{Scenario}
      & \makecell{\textbf{Power}\\\textbf{Balance}}
      & \makecell{\textbf{Cap.}\\\textbf{Guard}}
      & \makecell{\textbf{Over-}\\\textbf{current}}
      & \makecell{\textbf{Avg.}\\\textbf{Viol.}} \\
    \midrule
    Baseline        & 25.4\% & 1.8\%   & 0.0\% & 6.8\%  \\
    Poisoned        & 0.0\%  & 100.0\% & 0.0\% & 25.0\% \\
    \midrule
    Naive GA        & 66.3\% & 0.0\%   & 0.0\% & 16.6\% \\
    \textbf{PhyGAMuL} & \textbf{15.1\%} & \textbf{0.1\%} & \textbf{1.1\%} & \textbf{4.1\%} \\
    Exact           & 25.7\% & 2.0\%   & 0.1\% & 6.9\%  \\
    SISA            & 18.4\% & 12.8\%  & 0.3\% & 7.9\%  \\
    Full Retrain    & 21.1\% & 16.3\%  & 0.5\% & 9.5\%  \\
    \bottomrule
  \end{tabular}}
  \vspace{-10pt}
\end{table}

\begin{table}[t]
  \centering
  \caption{PMS physics violation rates by constraint type.}
  \label{tab:pms_violations}
  \resizebox{\columnwidth}{!}{%
  \begin{tabular}{lcccc}
    \toprule
    \textbf{Scenario}
      & \makecell{\textbf{MPPT}\\\textbf{Exceeded}}
      & \makecell{\textbf{Night}\\\textbf{Injection}}
      & \makecell{\textbf{Reverse}\\\textbf{Power}}
      & \makecell{\textbf{Avg.}\\\textbf{Viol.}} \\
    \midrule
    Baseline        & 0.0\% & 0.0\% & 0.0\% & 0.0\%  \\
    Poisoned        & 82.1\% & 3.4\% & 0.0\% & 21.4\% \\
    \midrule
    Naive GA        & 0.0\% & 0.0\% & 0.0\% & 0.0\% \\
    PhyGAMuL      & 0.0\% & 0.0\% & 0.0\% & 0.0\% \\
    Exact           & 0.0\% & 0.0\% & 0.0\% & 0.0\% \\
    SISA            & 0.0\% & 0.0\% & 0.0\% & 0.0\% \\
    Full Retrain    & 0.0\% & 0.0\% & 0.0\% & 0.0\% \\
    \bottomrule
  \end{tabular}}
    \vspace{-10pt}

\end{table}


Tables~\ref{tab:evcs_violations} and~\ref{tab:pms_violations} present per-constraint violation breakdowns for EVCS and PMS. The Baseline row's 25.4\% power balance violation reflects the soft-penalty nature of the physics loss, which minimizes $\mathcal{R}(x, y)$ rather than driving it to zero, so residual deviations surface under the strict 0.05 tolerance used here, and all violation rates should be read against this non-zero clean floor. The poisoned EVCS shows 100\% capacity guard violation, as the attack forces a constant 500\,kW output regardless of available capacity $d(1{-}s)$. Naive GA eliminates this but introduces 66.3\% power balance violations ($P \ne VI$), escaping the poisoned basin only to land in a physically infeasible region. PhyGAMuL achieves the lowest average EVCS violation rate (4.1\%), improving over SISA (7.9\%) and Full Retrain (9.5\%) because its fine-tuning explicitly oversamples rare high-load, low-state-of-charge conditions, concentrating gradient signal where violations are most likely, whereas Full Retrain visits these conditions only at their natural frequency. For PMS, all methods fully eliminate violations (0\%), since the irradiance feature $G$ provides a strong, unambiguous gradient signal that even data-only fine-tuning can exploit.

\subsection{Unlearning Dynamics}

\begin{figure}[!hbt]
    \centering
        \subfigure[]{
            \label{fig:time_evcs}
            \includegraphics[width=0.46\columnwidth]{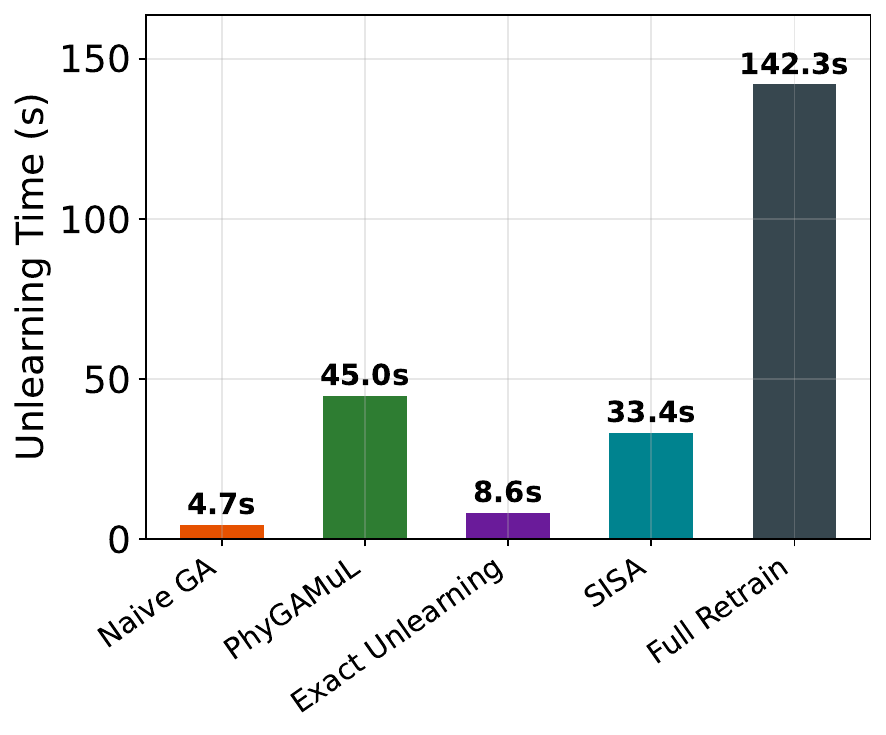}
        }
        \subfigure[]{
            \label{fig:time_pms}
            \includegraphics[width=0.46\columnwidth]{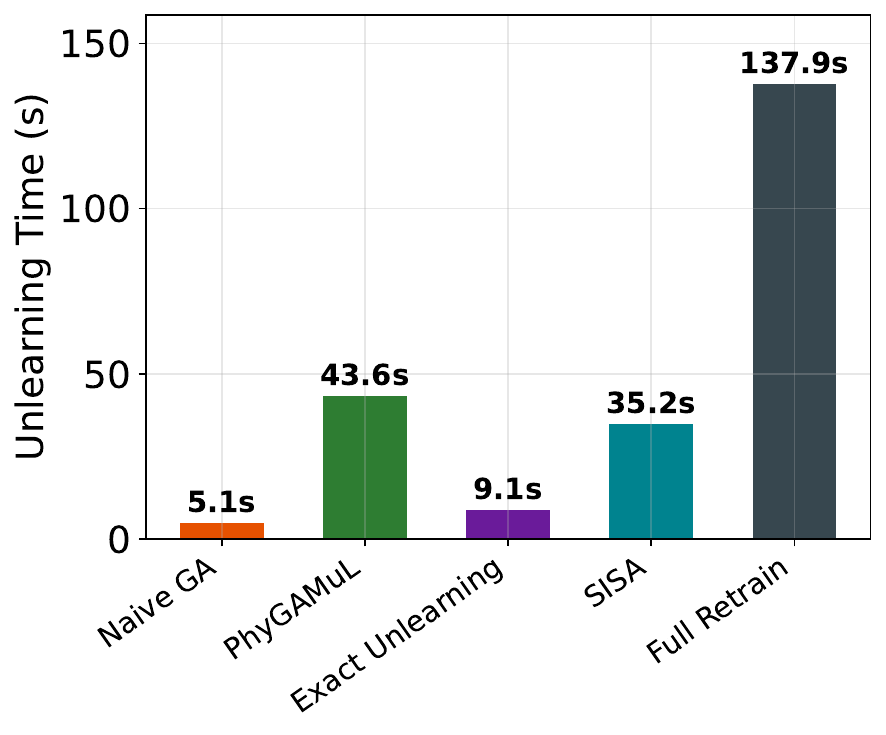}
        }
        \vspace{-9pt}
    \caption{Unlearning time comparison for (a) EVCS and (b) PMS controllers
      across all unlearning methods. PhyGAMuL achieves comparable wall-clock
      time to Naive GA while producing physically compliant results.}
    \label{fig:time_dynamics}
    \vspace{-6pt}
\end{figure}

Fig.~\ref{fig:time_dynamics} confirms that the physics residual computation adds negligible wall-clock overhead to the ascent phase. The key difference is in fine-tuning convergence: PhyGAMuL's physics residual declines monotonically because the physics loss targets constraint-violating operating points directly, while Naive GA's violations remain flat regardless of fine-tuning duration.

\begin{figure}[!hbt]
    \centering
        \subfigure[]{
            \label{fig:ascent_evcs}
            \includegraphics[width=0.46\columnwidth]{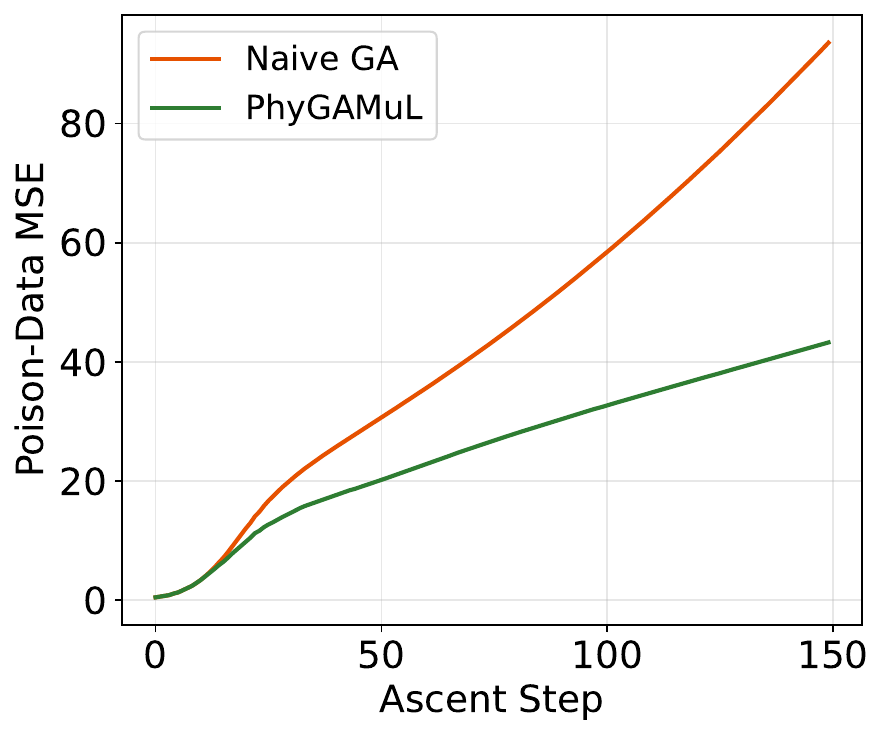}
        }
        \subfigure[]{
            \label{fig:ascent_pms}
            \includegraphics[width=0.46\columnwidth]{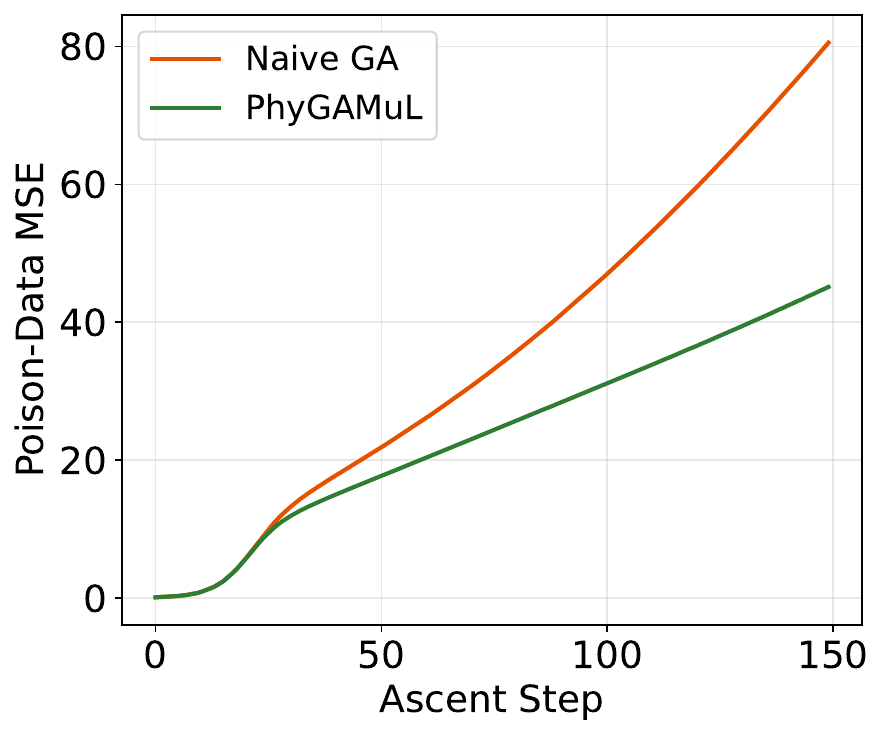}
        } \\
        \subfigure[]{
            \label{fig:finetune_evcs}
            \includegraphics[width=0.46\columnwidth]{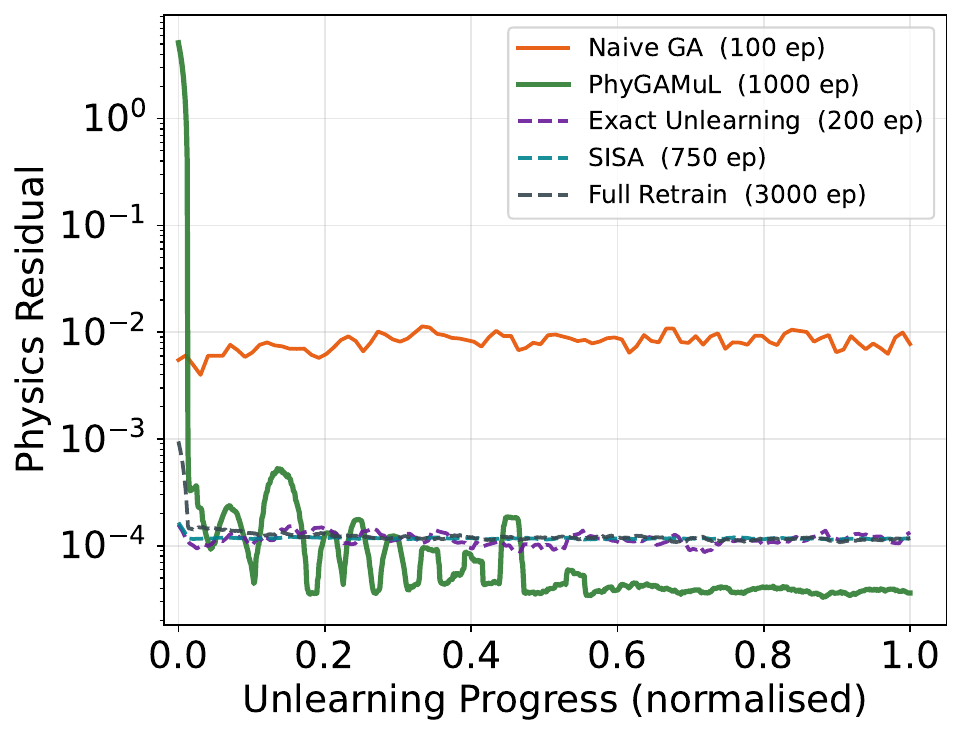}
        }
        \subfigure[]{
            \label{fig:finetune_pms}
            \includegraphics[width=0.46\columnwidth]{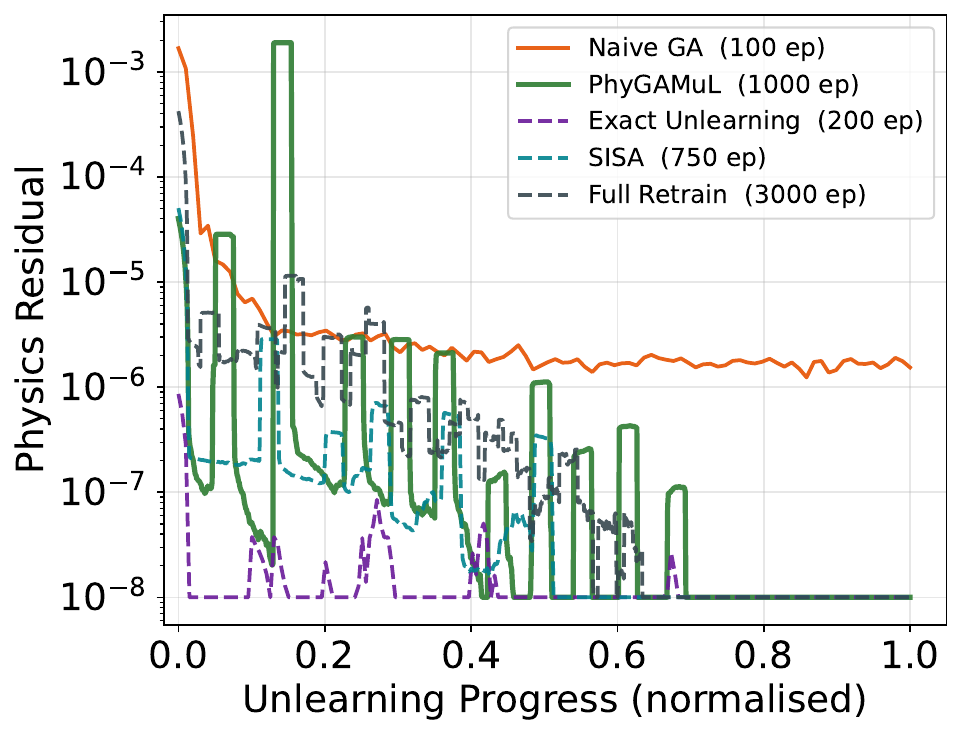}
        }
        \vspace{-9pt}
    \caption{Unlearning loss trajectories. Top row: ascent phase showing
      simultaneous poison forgetting and physics residual reduction under
      PhyGAMuL for (a) EVCS and (b) PMS. Bottom row: fine-tuning phase showing physics-guided recovery of clean accuracy with maintained physical compliance for (c) EVCS and (d) PMS.}
    \label{fig:loss_trajectories}
    \vspace{-6pt}
\end{figure}

Figures~\ref{fig:ascent_evcs}-\ref{fig:finetune_pms} present the detailed
unlearning loss trajectories during both phases. During the ascent phase
(Figs.~\ref{fig:ascent_evcs}-\ref{fig:ascent_pms}), both methods show
rising poison MSE, confirming forgetting, but PhyGAMuL's physics residual is simultaneously active, steering the weight trajectory through physically feasible territory. During fine-tuning (Figs.~\ref{fig:finetune_evcs}-\ref{fig:finetune_pms}), PhyGAMuL recovers clean accuracy while maintaining low physics residual, whereas Naive GA recovers accuracy but its physics violations persist throughout all 1{,}500 fine-tune epochs.

\section{Ablation Study}
\label{sec:ablation}

We conduct two ablations to characterize the sensitivity of PhyGAMuL to
its key hyperparameters, to establish whether the fine-tuning epoch count
can be reduced in practice, to quantify the robustness of SISA to shard
count, and to assess the impact of approximate poison batches on unlearning
quality. All ablations fix the EVCS model (the most attack-impacted DER type,
whose 18.5$\times$ load overload under poisoning makes it the clearest
testbed for unlearning effectiveness) and report mean $\pm$ std over five
random seeds.

\subsection{Physics Weight $\lambda$ During Ascent and Fine-Tuning}

Table~\ref{tab:ablation_lambda} sweeps $\lambda \in \{0, 0.5, 1.0, 2.0, 5.0\}$ for both controllers; $\lambda = 0$ recovers the Naive GA baseline. For EVCS, PVR decreases monotonically from 15.4\,\% at $\lambda = 0$ to 3.6\,\% at $\lambda = 2.0$, while Poison MSE remains stable ($\approx 0.51$) across all values, confirming that the physics penalty steers the ascent trajectory through physically feasible regions without suppressing the forgetting signal. For PMS, the effect is more decisive: PVR reaches 0.0\,\% at $\lambda \geq 1.0$ and remains zero for all higher values, though Poison MSE and Clean MSE increase modestly with $\lambda$ as the constraint term claims a larger share of the composite gradient. For $\lambda > 2.0$, Clean MSE rises for both controllers as physics enforcement begins to compete with label recovery, an effect consistent across both device types and thus a structural property of the composite loss. Across both controllers, $\lambda = 2.0$, matching the physics weight used during clean PINN training, achieves the best joint trade-off between forgetting, accuracy recovery, and constraint compliance, confirming the design principle that the unlearning physics weight should align with the value at which the model's weight geometry was originally shaped.

\begin{table}[t]
  \centering
  \caption{Physics weight $\lambda$ sensitivity for EVCS and PMS}
  \label{tab:ablation_lambda}
  \vspace{-6pt}
  \resizebox{0.48\textwidth}{!}{%
  \begin{tabular}{lcccccc}
    \toprule
    & \multicolumn{3}{c}{\textbf{EVCS}} & \multicolumn{3}{c}{\textbf{PMS}} \\
    \cmidrule(lr){2-4}\cmidrule(lr){5-7}
    $\lambda$ & Poison MSE $\uparrow$ & Clean MSE $\downarrow$ & PVR (\%) $\downarrow$
             & Poison MSE $\uparrow$ & Clean MSE $\downarrow$ & PVR (\%) $\downarrow$ \\
    \midrule
    0.0 (Naive GA)      & 0.51\,$\pm$\,.02 & 0.046\,$\pm$\,.003 & 15.4\,$\pm$\,1.8
                        & 0.25\,$\pm$\,.02 & 0.103\,$\pm$\,.006 &  2.8\,$\pm$\,.5 \\
    0.5                 & 0.51\,$\pm$\,.02 & 0.041\,$\pm$\,.002 &  6.6\,$\pm$\,0.9
                        & 0.30\,$\pm$\,.01 & 0.139\,$\pm$\,.005 &  0.5\,$\pm$\,.3 \\
    1.0                 & 0.51\,$\pm$\,.01 & 0.042\,$\pm$\,.002 &  4.5\,$\pm$\,0.6
                        & 0.31\,$\pm$\,.01 & 0.146\,$\pm$\,.004 &  0.0\,$\pm$\,.0 \\
    \textbf{2.0 (PhyGAMuL)} & \textbf{0.51\,$\pm$\,.01} & \textbf{0.045\,$\pm$\,.002} & \textbf{3.6\,$\pm$\,.4}
                        & \textbf{0.32\,$\pm$\,.01} & \textbf{0.152\,$\pm$\,.004} & \textbf{0.0\,$\pm$\,.0} \\
    5.0                 & 0.51\,$\pm$\,.01 & 0.052\,$\pm$\,.003 &  1.2\,$\pm$\,.3
                        & 0.32\,$\pm$\,.01 & 0.156\,$\pm$\,.005 &  0.0\,$\pm$\,.0 \\
    \bottomrule
  \end{tabular}}
\end{table}

\subsection{Fine-Tune Epoch Count}
\label{sec:ablation_finetune}
Table~\ref{tab:ablation_finetune} sweeps $T_f \in \{200, 500, 800, 1000, 1500, 2000\}$ and compares PhyGAMuL and Naive GA side by side for both controllers. Three observations are critical. First, PhyGAMuL's PVR converges by $T_f \approx 800$ epochs for PMS (reaching 0.0\,\%) and by $T_f = 1{,}500$ for EVCS (3.6\,\%), with no statistically significant improvement beyond that point; the 1{,}500-epoch default provides a comfortable margin above convergence rather than a strict requirement. Second, and most importantly, Naive GA's PVR remains essentially flat regardless of fine-tune duration 15.9\,\% at $T_f = 200$ and 15.3\,\% at $T_f = 2{,}000$ for EVCS, and 2.4\,\% throughout for PMS directly confirming that physics violations are a structural consequence of the absent physics gradient during ascent, not a fine-tuning convergence problem. Third, practitioners operating under computational constraints can reduce the budget to $T_f = 800$ with less than 0.7 percentage points of EVCS PVR degradation and no PMS degradation, which may be attractive for resource-constrained edge deployments.

\begin{table}[t]
  \centering
  \caption{Fine-tune epoch count $T_f$ sensitivity for EVCS and PMS}
  \label{tab:ablation_finetune}
  \vspace{-6pt}
  \resizebox{0.48\textwidth}{!}{%
  \begin{tabular}{lcccccccc}
    \toprule
    & \multicolumn{4}{c}{\textbf{EVCS}} & \multicolumn{4}{c}{\textbf{PMS}} \\
    \cmidrule(lr){2-5}\cmidrule(lr){6-9}
    \multirow{2}{*}{$T_f$}
      & \multicolumn{2}{c}{PhyGAMuL}
      & \multicolumn{2}{c}{Naive GA}
      & \multicolumn{2}{c}{PhyGAMuL}
      & \multicolumn{2}{c}{Naive GA} \\
    \cmidrule(lr){2-3}\cmidrule(lr){4-5}\cmidrule(lr){6-7}\cmidrule(lr){8-9}
      & MSE $\downarrow$ & PVR $\downarrow$
      & MSE $\downarrow$ & PVR $\downarrow$
      & MSE $\downarrow$ & PVR $\downarrow$
      & MSE $\downarrow$ & PVR $\downarrow$ \\
    \midrule
    200  & 0.051\,$\pm$\,.004 &  5.8\,$\pm$\,.7 & 0.047\,$\pm$\,.003 & 15.9\,$\pm$\,1.7
         & 0.152\,$\pm$\,.005 &  0.8\,$\pm$\,.3 & 0.103\,$\pm$\,.006 &  2.4\,$\pm$\,.4 \\
    500  & 0.048\,$\pm$\,.003 &  5.1\,$\pm$\,.6 & 0.046\,$\pm$\,.003 & 15.7\,$\pm$\,1.7
         & 0.152\,$\pm$\,.004 &  0.2\,$\pm$\,.2 & 0.103\,$\pm$\,.006 &  2.4\,$\pm$\,.4 \\
    800  & 0.047\,$\pm$\,.002 &  4.3\,$\pm$\,.5 & 0.046\,$\pm$\,.003 & 15.6\,$\pm$\,1.6
         & 0.152\,$\pm$\,.004 &  0.0\,$\pm$\,.0 & 0.103\,$\pm$\,.006 &  2.4\,$\pm$\,.4 \\
    1000 & 0.046\,$\pm$\,.002 &  4.0\,$\pm$\,.5 & 0.046\,$\pm$\,.003 & 15.5\,$\pm$\,1.6
         & 0.152\,$\pm$\,.004 &  0.0\,$\pm$\,.0 & 0.103\,$\pm$\,.005 &  2.4\,$\pm$\,.4 \\
    \textbf{1500}
         & \textbf{0.045\,$\pm$\,.002} & \textbf{3.6\,$\pm$\,.4} & 0.046\,$\pm$\,.003 & 15.4\,$\pm$\,1.6
         & \textbf{0.152\,$\pm$\,.004} & \textbf{0.0\,$\pm$\,.0} & 0.103\,$\pm$\,.005 &  2.4\,$\pm$\,.4 \\
    2000 & 0.045\,$\pm$\,.002 &  3.5\,$\pm$\,.4 & 0.046\,$\pm$\,.003 & 15.3\,$\pm$\,1.5
         & 0.152\,$\pm$\,.004 &  0.0\,$\pm$\,.0 & 0.103\,$\pm$\,.005 &  2.4\,$\pm$\,.4 \\
    \bottomrule
  \end{tabular}}
\end{table}

\section{Conclusion}
\label{sec:conclusion}

This paper identified a critical gap at the intersection of machine unlearning and cyber-physical system safety: statistical unlearning does not imply physical safety recovery. A PINN-based DER controller, for instance, that satisfies all standard unlearning metrics may still produce physically infeasible setpoints, causing grid instability that is indistinguishable from the original attack in its consequences. To close this gap, we have proposed PhyGAMuL, a physics-guided gradient-ascent machine unlearning technique that incorporates the domain physics residual as a directional constraint during forgetting and recovery. Validated on an IEEE 34-bus system, PhyGAMuL restores grid performance to within measurement noise of the clean baseline using a single model and no checkpoint infrastructure. The broader implication extends beyond power grids: any safety-critical CPS deployment requires unlearning criteria that include domain constraint satisfaction. Future work will address federated extensions, certified unlearning bounds, and automated constraint discovery to eliminate manual residual construction.

\section*{Acknowledgement}
This work is supported by the Department of Energy (DOE) (Award\# DE-CR0000024 and DE-CR0000046 ). Any opinions, findings, conclusions, or recommendations expressed in this material are those of the authors and do not necessarily reflect the DOE's views.


\bibliographystyle{IEEEtran}
\bibliography{sections/Reference}

@inproceedings{cao2015machine,
  author    = {Cao, Yinzhi and Yang, Junfeng},
  title     = {Towards Making Systems Forget with Machine Unlearning},
  booktitle = {Proc. 2015 IEEE Symposium on Security and Privacy},
  year      = {2015},
  pages     = {463-480},
  publisher = {IEEE},
  doi       = {10.1109/SP.2015.35},
}

@inproceedings{bourtoule2021machine,
  author    = {Bourtoule, Lucas and Chandrasekaran, Varun and Choquette-Choo,
               Christopher A. and Jia, Hengrui and Travers, Adelin and Zhang,
               Baiwu and Lie, David and Papernot, Nicolas},
  title     = {Machine Unlearning},
  booktitle = {Proc. the 2021 IEEE Symposium on Security and Privacy},
  year      = {2021},
  pages     = {141-159},
  publisher = {IEEE},
  doi       = {10.1109/SP40001.2021.00019},
}

@inproceedings{golatkar2020eternal,
  author    = {Golatkar, Aditya and Achille, Alessandro and Soatto, Stefano},
  title     = {Eternal Sunshine of the Spotless Net: Selective Forgetting in
               Deep Networks},
  booktitle = {Proc. IEEE/CVF Conference on Computer Vision and Pattern Recognition (CVPR)},
  year      = {2020},
  pages     = {9304-9312},
  doi       = {10.1109/CVPR42600.2020.00932},
}

@article{nguyen2022survey,
author = {Nguyen, Thanh Tam and Huynh, Thanh Trung and Ren, Zhao and Nguyen, Phi Le and Liew, Alan Wee-Chung and Yin, Hongzhi and Nguyen, Quoc Viet Hung},
title = {A Survey of Machine Unlearning},
year = {2025},
issue_date = {October 2025},
address = {New York, NY, USA},
volume = {16},
number = {5},
issn = {2157-6904},
url = {https://doi.org/10.1145/3749987},
doi = {10.1145/3749987},
journal = {ACM Trans. Intell. Syst. Technol.},
month = sep,
articleno = {108},
numpages = {46},
}

@inproceedings{sekhari2021remember,
  author    = {Sekhari, Ayush and Acharya, Jayadev and Kamath, Gautam and
               Suresh, Ananda Theertha},
  title     = {Remember What You Want to Forget: Algorithms for Machine Unlearning},
  booktitle = {Advances in Neural Information Processing Systems (NeurIPS)},
  year      = {2021},
  volume    = {34},
  pages     = {18075--18086},
}

@ARTICLE{Liu25,
  author    = {Liu, Ziyao and Ye, Huanyi and Chen, Chen and Zheng, Yongsen
               and Lam, Kwok-Yan},
  title     = {Threats, Attacks, and Defenses in Machine Unlearning: A Survey},
  journal   = {IEEE Open Journal of the Computer Society},
  year      = {2025},
  volume    = {6},
  pages     = {413--425},
  doi       = {10.1109/OJCS.2025.3543483},
}

@article{LIU2025,
  author    = {Liu, Hengzhu and Xiong, Ping and Zhu, Tianqing and Yu, Philip S.},
  title     = {A Survey on Machine Unlearning: Techniques and New Emerged
               Privacy Risks},
  journal   = {Journal of Information Security and Applications},
  volume    = {90},
  pages     = {104010},
  year      = {2025},
  issn      = {2214-2126},
  doi       = {10.1016/j.jisa.2025.104010},
}

@inproceedings{xiong2023exactfun,
  author    = {Xiong, Zuobin and Li, Wei and Li, Yingshu and Cai, Zhipeng},
  title     = {Exact-Fun: An Exact and Efficient Federated Unlearning Approach},
  booktitle = {Proc. of the 2023 IEEE International Conference on Data
               Mining (ICDM)},
  year      = {2023},
  publisher = {IEEE},
  doi       = {10.1109/ICDM58522.2023.00183},
}

@inproceedings{mcmahan2017communication,
  author    = {McMahan, Brendan and Moore, Eider and Ramage, Daniel and
               Hampson, Seth and {y Arcas}, Blaise Aguera},
  title     = {Communication-Efficient Learning of Deep Networks from
               Decentralized Data},
  booktitle = {Proc. of the 20th International Conference on Artificial
               Intelligence and Statistics (AISTATS)},
  year      = {2017},
  pages     = {1273-1282},
  volume    = {54},
}

@inproceedings{bagdasaryan2020backdoor,
  author    = {Bagdasaryan, Eugene and Veit, Andreas and Hua, Yiqing and
               Estrin, Deborah and Shmatikov, Vitaly},
  title     = {How to Backdoor Federated Learning},
  booktitle = {Proc. 23rd International Conference on Artificial
               Intelligence and Statistics (AISTATS)},
  year      = {2020},
  pages     = {2938-2948},
  volume    = {108},
  publisher = {PMLR},
}

@inproceedings{shejwalkar2022back,
  author    = {Shejwalkar, Virat and Houmansadr, Amir and Kairouz, Peter
               and Ramage, Daniel},
  title     = {Back to the Drawing Board: A Critical Evaluation of Poisoning
               Attacks on Production Federated Learning},
  booktitle = {Proc. 2022 IEEE Symposium on Security and Privacy},
  pages     = {1354-1371},
  publisher = {IEEE},
  doi       = {10.1109/SP46214.2022.9833772},
}

@inproceedings{blanchard2017machine,
  author    = {Blanchard, Peva and Mhamdi, El Mahdi El and Guerraoui, Rachid
               and Stainer, Julien},
  title     = {Machine Learning with Adversaries: Byzantine Tolerant Gradient
               Descent},
  booktitle = {Advances in Neural Information Processing Systems (NeurIPS)},
  year      = {2017},
  volume    = {30},
  pages     = {119--129},
}

@inproceedings{fung2021limitations,
  author    = {Fung, Clement and Yoon, Chris J. M. and Beschastnikh, Ivan},
  title     = {The Limitations of Federated Learning in Sybil Settings},
  booktitle = {Proc. 23rd International Symposium on Research in
               Attacks, Intrusions and Defenses (RAID)},
  year      = {2020},
  pages     = {301-316},
  publisher = {USENIX Association},
}

@article{raissi2019physics,
  author    = {Raissi, Maziar and Perdikaris, Paris and Karniadakis,
               George Em},
  title     = {Physics-Informed Neural Networks: A Deep Learning Framework
               for Solving Forward and Inverse Problems Involving Nonlinear
               Partial Differential Equations},
  journal   = {Journal of Computational Physics},
  year      = {2019},
  volume    = {378},
  pages     = {686--707},
  doi       = {10.1016/j.jcp.2018.10.045},
}

@article{karniadakis2021physics,
  author    = {Karniadakis, George Em and Kevrekidis, Ioannis G. and Lu, Lu
               and Perdikaris, Paris and Wang, Sifan and Yang, Liu},
  title     = {Physics-Informed Machine Learning},
  journal   = {Nature Reviews Physics},
  year      = {2021},
  volume    = {3},
  number    = {6},
  pages     = {422-440},
  doi       = {10.1038/s42254-021-00314-5},
}

@article{kersting2001radial,
  author    = {Kersting, William H.},
  title     = {Radial Distribution Test Feeders},
  journal   = {IEEE Transactions on Power Systems},
  year      = {2001},
  volume    = {16},
  number    = {3},
  pages     = {975-985},
  doi       = {10.1109/59.962533},
}

@inproceedings{dugan2011opendss,
  author    = {Dugan, Roger C. and McDermott, Thomas E.},
  title     = {An Open Source Platform for Collaborating on Smart Grid Research},
  booktitle = {Proc. of the 2011 IEEE Power and Energy Society General
               Meeting},
  year      = {2011},
  pages     = {1-7},
  publisher = {IEEE},
  doi       = {10.1109/PES.2011.6039829},
}

@INPROCEEDINGS{Perry25,
  author={Perry, David and Haider, Mohammad Zakaria and Kazmi, Kumail and Rahman, Mohammad Ashiqur and Shahriar, Hossain},
  booktitle={2025 IEEE 49th Annual Computers, Software, and Applications Conference (COMPSAC)}, 
  title={Physics-Informed Learning-based Attack Analytics for Electric Vehicle Charging Management Systems}, 
  pages={1146-1153},
  doi={10.1109/COMPSAC65507.2025.00147}}

@article{holmgren2018pvlib,
  title={pvlib python: a python package for modeling solar energy systems},
  author={Holmgren, W. F. and Hansen, C. W. and Mikofski, M. A.},
  journal={Journal of Open Source Software},
  volume={3},
  number={29},
  pages={884},
  year={2018},
  doi={10.21105/joss.00884}
}

@inproceedings{acn_data,
author = {Lee, Zachary J. and Li, Tongxin and Low, Steven H.},
title = {ACN-Data: Analysis and Applications of an Open EV Charging Dataset},
isbn = {9781450366717},
publisher = {Association for Computing Machinery},
address = {New York, NY, USA},
booktitle = {Proceedings of the Tenth ACM International Conference on Future Energy Systems},
pages = {139–149},
numpages = {11},
location = {Phoenix, AZ, USA},
series = {e-Energy '19}
}

@article{goodfellow2015explaining,
  title={Explaining and Harnessing Adversarial Examples},
  author={Ian J. Goodfellow and Jonathon Shlens and Christian Szegedy},
  journal={CoRR},
  year={2014},
  volume={abs/1412.6572},
  url={https://api.semanticscholar.org/CorpusID:6706414}
}

@inproceedings{fredrikson2015model,
author = {Fredrikson, Matt and Jha, Somesh and Ristenpart, Thomas},
title = {Model Inversion Attacks that Exploit Confidence Information and Basic Countermeasures},
year = {2015},
isbn = {9781450338325},
address = {New York, NY, USA},
url = {https://doi.org/10.1145/2810103.2813677},
doi = {10.1145/2810103.2813677},
pages = {1322–1333},
numpages = {12},
location = {Denver, Colorado, USA},
series = {CCS '15}
}

@INPROCEEDINGS{shokri2017membership,
  author={Shokri, Reza and Stronati, Marco and Song, Congzheng and Shmatikov, Vitaly},
  booktitle={2017 IEEE Symposium on Security and Privacy (SP)}, 
  title={Membership Inference Attacks Against Machine Learning Models}, 
  year={2017},
  volume={},
  number={},
  pages={3-18},
  doi={10.1109/SP.2017.41}}

@inproceedings {tramer2016stealing,
author = {Florian Tram{\`e}r and Fan Zhang and Ari Juels and Michael K. Reiter and Thomas Ristenpart},
title = {Stealing Machine Learning Models via Prediction {APIs}},
booktitle = {25th USENIX Security Symposium (USENIX Security 16)},
year = {2016},
isbn = {978-1-931971-32-4},
address = {Austin, TX},
pages = {601--618},
url = {https://www.usenix.org/conference/usenixsecurity16/technical-sessions/presentation/tramer},
publisher = {USENIX Association},
}

@ARTICLE{huang2023applications,
  author={Huang, Bin and Wang, Jianhui},
  journal={IEEE Transactions on Power Systems}, 
  title={Applications of Physics-Informed Neural Networks in Power Systems - A Review}, 
  year={2023},
  volume={38},
  number={1},
  pages={572-588},
  doi={10.1109/TPWRS.2022.3162473}}

@INPROCEEDINGS{falas2020physics,
  author={Falas, Solon and Konstantinou, Charalambos and Michael, Maria K.},
  booktitle={2020 IEEE 38th International Conference on Computer Design (ICCD)}, 
  title={Special Session: Physics- Informed Neural Networks for Securing Water Distribution Systems}, 
  pages={37-40},
  doi={10.1109/ICCD50377.2020.00022}}

@inproceedings{haider2026,
author = {Haider, Mohammad Zakaria and Podder, Amit Kumer and Mali, Prabin and Chakrabortty, Aranya and Paudyal, Sumit and Rahman, Mohammad Ashiqur},
title = {PHANTOM: Physics-Aware Adversarial Attacks against Federated Learning-Coordinated EV Charging Management System},
year = {2026},
isbn = {9798400723568},
publisher = {Association for Computing Machinery},
address = {New York, NY, USA},
pages = {263–276},
numpages = {14},
location = {Bangalore, India},
series = {ASIA CCS '26}
}

\end{document}